\documentclass[twoside,leqno,twocolumn]{article}

\usepackage[letterpaper]{geometry}
\usepackage{cite}
\usepackage{ltexpprt}
\usepackage{hyperref}
\usepackage{multicol}
\usepackage{graphicx}
\usepackage{multirow}
\usepackage{booktabs}
\usepackage{amssymb}
\usepackage{bbm}
\begin{document}

\newcommand\relatedversion{}

\title{A Kalman Filter Based Framework for Monitoring the Performance of In-Hospital Mortality Prediction Models Over Time}
\author{Jiacheng Liu\thanks{Dept. of Computer Science, University of Minnesota.}
\and Lisa Kirkland\thanks{Abbott Northwestern Hospital, Allina Health}
\and Jaideep Srivastava\thanks{Dept. of Computer Science, University of Minnesota.}}

\date{}

\maketitle

% Copyright Statement
% When submitting your final paper to a SIAM proceedings, it is requested that you include
% the appropriate copyright in the footer of the paper.  The copyright added should be
% consistent with the copyright selected on the copyright form submitted with the paper.
% Please note that "20XX" should be changed to the year of the meeting.

% Default Copyright Statement
\fancyfoot[R]{\scriptsize{Copyright \textcopyright\ 2023 by SIAM\\
Unauthorized reproduction of this article is prohibited}}

% Depending on which copyright you agree to when you sign the copyright form, the copyright
% can be changed to one of the following after commenting out the default copyright statement
% above.

%\fancyfoot[R]{\scriptsize{Copyright \textcopyright\ 20XX\\
%Copyright for this paper is retained by authors}}

%\fancyfoot[R]{\scriptsize{Copyright \textcopyright\ 20XX\\
%Copyright retained by principal author's organization}}

%\pagenumbering{arabic}
%\setcounter{page}{1}%Leave this line commented out.

\begin{abstract} \small\baselineskip=9pt Unlike in a clinical trial, where researchers get to determine the least number of positive and negative samples required, or in a machine learning study where the size and the class distribution of the validation set is static and known, in a real-world scenario, there is little control over the size and distribution of incoming patients. As a result, when measured during different time periods, evaluation metrics like Area under the Receiver Operating Curve (AUCROC) and Area Under the Precision-Recall Curve(AUCPR) may not be directly comparable. Therefore, in this study, for binary classifiers running in a long time period, we proposed to adjust these performance metrics for sample size and class distribution, so that a fair comparison can be made between two time periods. Note that the number of samples and the class distribution, namely the ratio of positive samples, are two robustness factors which affect the variance of AUCROC. To better estimate the mean of performance metrics and understand the change of performance over time, we propose a Kalman filter based framework with extrapolated variance adjusted for the total number of samples and the number of positive samples during different time periods. The efficacy of this method is demonstrated first on a synthetic dataset and then retrospectively applied to a 2-days ahead in-hospital mortality prediction model for COVID-19 patients during 2021 and 2022. Further, we conclude that our prediction model is not significantly affected by the evolution of the disease, improved treatments and changes in hospital operational plans.\end{abstract}

\section{Introduction.}
Area under the Receiver Operating Curve (AUCROC) is widely used as an evaluation metric of predictive models with binary outcomes. In health informatics, such prediction targets can be diagnosis of a particular disease, malignancy of a tumor, risk of ICU transfers and risk of in-hospital mortality. A typical research workflow involves derivation and training of the predictive model, then the model will be evaluated on a held-out test dataset. Unless the model is deployed, no further performance metrics (such as AUCROC) will be recorded.

Continuous monitoring is essential for any predictive model to be operationalized, so that adjustments of model parameters (such as decision thresholds), and decisions of whether the model is outdated can be properly and timely made. Unlike in a controlled environment where the performance is either evaluated on desired class distribution and sample size (i.e., in a clinical trial), or reported only once (i.e., in a machine learning study), tracking model performance over time requires multiple and regular tests of the model. This brings challenges in an evolving environment because the size and the class distribution (namely the ratio of positive class, if binary prediction target assumed) of the incoming data batch is no longer the same as of the initial training/validation dataset. To be more specific, the number of samples and the class distribution, are two robustness factors which affects the variance of performance metrics like AUCROC. Besides, a robustness factor for one performance evaluation metric can be a dominant factor of another evaluation metric. For example, number of ground truth positive samples only affects the variance of AUCROC, but both mean and variance of Area under the Precision-Recall Curve (AUCPR).

%Root cause analysis
\begin{figure*}[t]
  \centering
  \includegraphics[width=14cm]{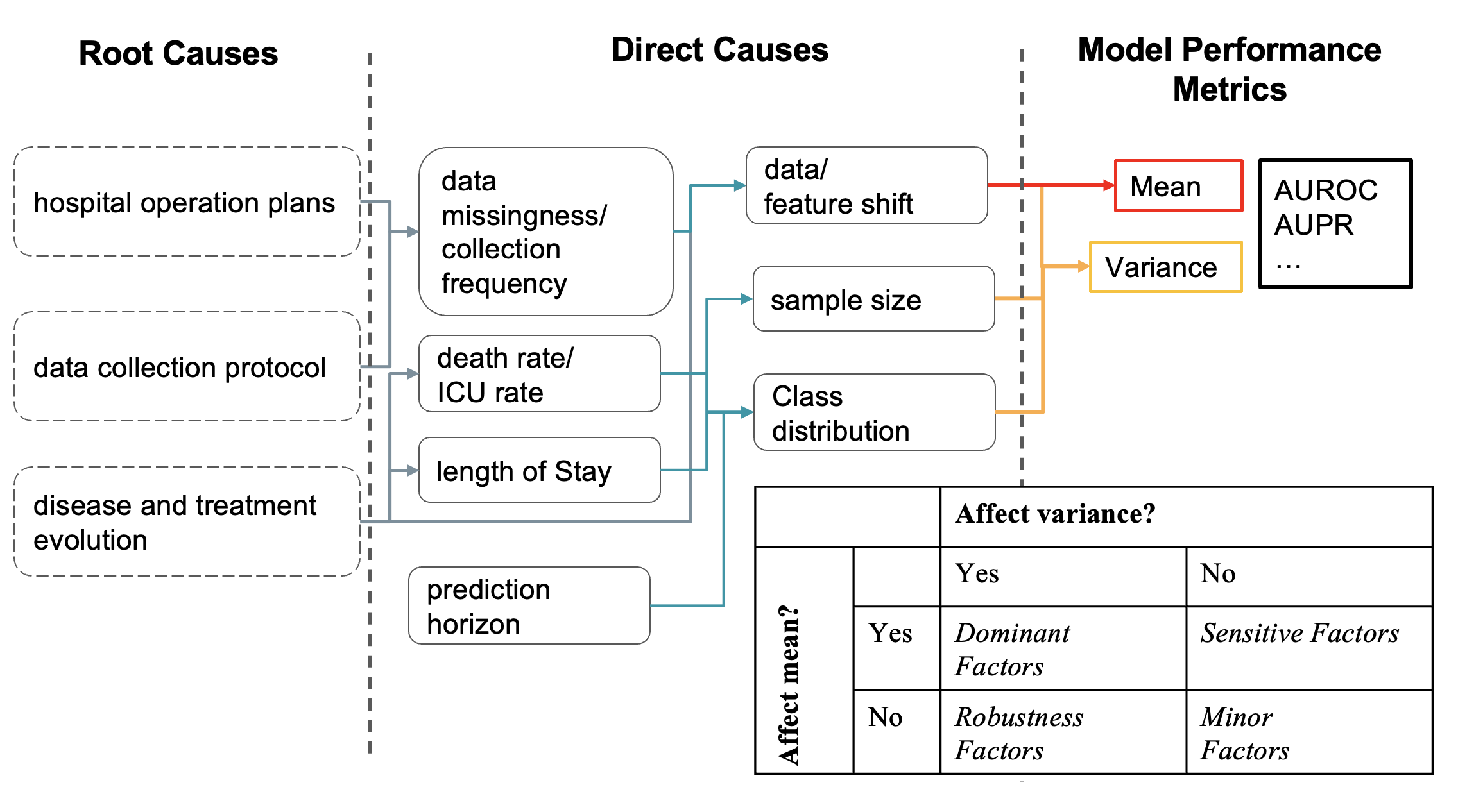}
  \caption{\centering Causes of change in predictive model performance over time. The direct causes are those factors that affect mean and variance of evaluation metrics. Note that all factors in this figure may change over time. Besides, the categorization of factor types is metric dependent. Number of ground truth positive samples only affects the variance of AUCROC, but both mean and variance of Area under the Precision-Recall Curve.}
  \label{fig:causes}
\end{figure*}

In this paper, we focus on the problem of tracking the mean AUCROC of binary classifiers over time in an environment of changing sample size and positive ratio. Since small and imbalanced datasets are quite common in the healthcare domain, bootstrapping method will not always work. Therefore, a one dimensional Kalman filter based framework is proposed, where a simple constant dynamic is employed, and the variance of the next time step is extrapolated in a sample size/positive ratio adjusted way. Furthermore, upon the appearance of extremely skewed class distribution, e.g., only 10 positive case and 490 negative cases, we proposed to use variance upper bound, which is adjusted for the sample size and positive ratio, instead of the sample variance inferred from the current test data batch. We argue that the number of positive and negative cases must be taken into account as dominant, robustness or sensitive factors of performance change, as they may swing a lot in a real world scenario. Therefore, a Kalman filter based framework is proposed. The framework is flexible enough to incorporate with different evaluation metrics under different assumptions of sample size and class distributions. Our contributions can be summarized as the following.
\begin{itemize}
    \item A layered model for performance change analysis. We point out that the number of positive and negative samples must be considered when explaining the change of model performance, as illustrated in Fig \ref{fig:causes}.
    \item A Kalman Filter based framework for estimating model performance over time. Following the analyses of dominant factors, sensitive factors, and robustness factors, a Kalman filter is a natural choice when the combined effects of these factors weigh in (Fig. 3.). Further, we provide rationales to use the variance upper bound instead of the sample variance, if the number of positive cases is extremely low. (Table 2.)
    \item Retrospective filtered performance of 2 days ahead in-hospital mortality prediction model for COVID-19 patients. The proposed algorithm is applied to a set of COVID-19 patients admitted between 2020 June and 2022 December. The model is trained on 2020 data. Then we report the test performance on the year of 2021 and 2022. The result suggests a consistently high prediction performance.
\end{itemize}

The rest of the paper is organized in this way. Section \ref{sec:2} gives the context of the in-hospital mortality prediction problem where the topic of this paper, the problem of monitoring prediction model performance arises. Section \ref{sec:rw} summarizes related work. We describe our proposed method in section \ref{sec:method}, starting from DeLong's method for estimating the variance of AUCROC. The questions raised in section \ref{sec:2} are answered by experiments on both synthetic datasets and the real world COVID-19 patients dataset. In the discussion section\ref{sec:discussion} we note the cons and pros, and highlight an interesting finding.

\section{Motivation}\label{sec:2}
\begin{figure}[h]
  \centering
  \includegraphics[width=\linewidth]{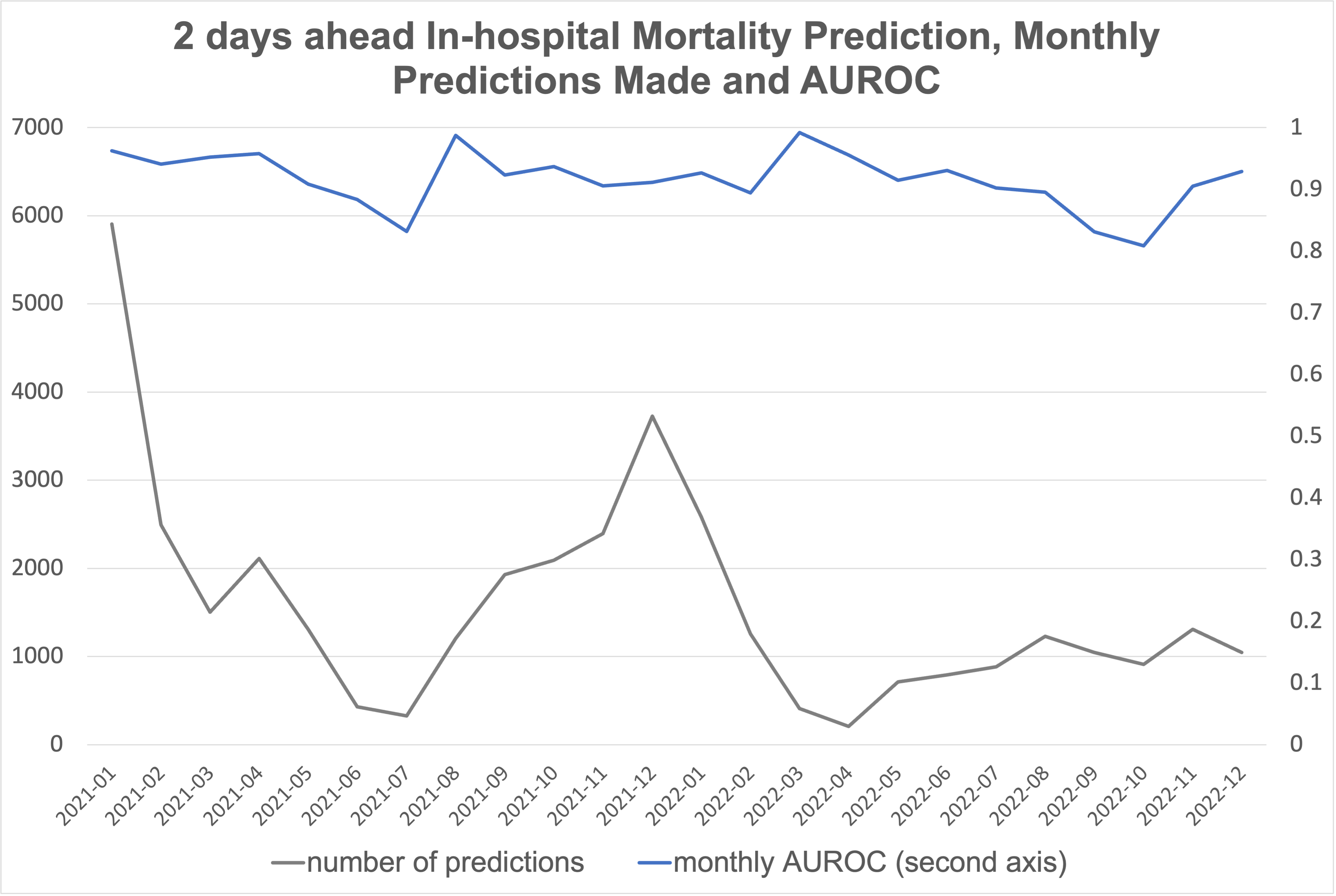}
  \caption{\centering Monthly AUCROC performance (the blue line) of 2 days ahead in-hospital mortality prediction model for COVID-19 patients. The model is trained on 2020 data only then tested retrospectively in 2021 and 2022. The grey line suggests a strong seasonal trend of number of hospitalized patients per month. Generally, the scale of performance fluctuation, in another word, sample variance, tends to be large when number of patients is low. This is no coincident.}
  \label{fig:monthlyAUCROC}
\end{figure}

The direct motivation of this study is demonstrated in Fig. \ref{fig:monthlyAUCROC}, where the blue line is AUCROC, calculated monthly for a 2 days ahead COVID-19 in-hospital mortality binary classifier during 2021 and 2022. The prediction model scores the hospitalized COVID-19 patients daily, indicating the risk of mortality in the next two days. The prediction model is trained only based on data in year 2020, and has not been retrained afterwards. The grey line is the total number of predictions made in that month. These are COVID-19 patients admitted to general wards or Intensive Care Unit(ICU) of Abbott Northwestern Hospital at Minneapolis from 2020 to 2022. The number of predictions shows an obvious seasonal trend as expected, as there are fewer patients in Spring and Summer. Basic statistics about the dataset are given in table \ref{tab:data statistics}. Please refer to supplementary materials for more details such as inclusion/exclusion criteria and summary statistics of demographics and vital signs. This section also provides a brief background of the binary prediction problem.

To the point of this paper, it is worth noting that the valleys of grey line correspond to the period when big fluctuations happens in the blue line. Recall the analysis made in Fig. \ref{fig:causes}, this is likely no coincidence. Given the fact that the 2 days ahead COVID-19 in-hospital mortality binary classifier is trained solely on 2020 data, we are curious to know, if any of the changes in disease variants\cite{mohammadi2021impact}, treatment approaches\cite{stasi2020treatment, pourkarim2022molnupiravir}, vaccination status\cite{andreadakis2020covid}, has impact on the performance of the model. These questions will be answered by the proposed method later in section \ref{sec:exp}. Also notice that, the number of predictions made in each month is affected by the number of incoming patients and the average length of hospital stay at the same time, e.g., same number of patients with shorter length of stays, there will be less number of predictions per month in total.

\subsection{2 days ahead In-hospital Mortality Prediction for COVID-19 Patients}

Models are trained to predict risk of in-hospital mortality in the next 2 days for both non-ICU (referred as ``floor'' patients) and ICU patients. The prediction horizon is determined to be 2 days since 1 day ahead prediction suffers the most from imbalanced labels and it leaves little intervention time for physicians. While the end-of-stay prediction task aims to characterize the overall risk during the entire stay, it provides little clues to the imminent danger of deterioration. According to the data selection criteria mentioned in the supplementary materials, all patient stays are at least 2 days long.

Table \ref{tab:data statistics} has the summary statistics of the dataset, stratified by both years and ICU status. If a patient is transferred to the ICU at least once during the entire hospital stay, it will count as a patient in the ICU stratum. In terms of end-of-stay outcomes and lengths of stay, while there is a clear distinction between the floor patients and ICU patients, the annual change is minimal. The dataset includes 7,080 patients from 2020 to 2022 with an average death rate of 10.95 \% and ICU rate of 27.2 \%. Please refer to the supplementary materials for more details about the summary statistics. More details about this dataset can be found in section \ref{sec:exp} and supplementary materials.
\begin{table*}
    \centering
        \begin{tabular}{rrrrrrr}
        \toprule
        & \multicolumn{2}{c}{\textbf{2020}} & \multicolumn{2}{c}{\textbf{2021}}& \multicolumn{2}{c}{\textbf{2022}} \\
        & Floor & ICU& Floor & ICU& Floor &  ICU \\
        \hline
        \midrule
\textbf{\# patients} & 2915(78.81\%) & 784(21.19\%) & 1551(72.48\%) & 589(27.52\%) & 533(51.90\%) & 494(48.10\%) \\
\textbf{\# deaths}   & 172(5.90\%) & 256(32.65\%) & 44(5.90\%)  & 192(32.65\%) & 13(5.90\%)  & 75(32.65\%) \\
\textbf{Length of Stay} &   & &  & & &  \\
mean & 8.17 &  18.14  &  8.16   &  17.55   &  6.93  & 12.16   \\
median& 6 & 14 & 6  &  13  &  5   &  8  \\
std & 6.95 &  16.76  & 6.28 & 13.79 \\
         \bottomrule
        \end{tabular}
        \caption{Summary statistics of COVID-19 dataset, stratified by year} 
        \label{tab:data statistics}
\end{table*}

\section{Related Work} \label{sec:rw}

% AUCROC, bootstrapping
Area Under the Receiver Operator Curve(AUCROC) is a common evaluation metric in biostatistics\cite{hillis2008recent,liu2013roc} and machine learning studies\cite{jiao2016performance}. Following the Mann-Whitney estimator of AUCROC \ref{eq:1}, there are two main methods for estimating the variance of AUCROC, Obuchowski's method\cite{obuchowski1997sample} and DeLong's method\cite{delong1988comparing,sun2014fast}. Bootstrapping is another commonly used empirical estimator of variance. However, in the case of few samples available, the performance of which quickly deteriorates\cite{hanley1988robustness, airola2009comparison,poynard2007variability}. Kalman filters are widely applied to various areas\cite{galanis2002one,meinhold1983understanding}. \cite{welch1995introduction} gives a nice introduction to Kalman filters. 

% covid ihm
Lots of research efforts have been devoted to triage or predict risk in-hospital mortality for COVID-19 patients during the last 3 years\cite{bachtiger2020machine}. In the area of predictive modeling, there are studies based on time-varying variables\cite{wu2020risk,tello2022machine} and images\cite{baig2020evidence}. The algorithms range from non-deep methods like XGBoost\cite{wu2020risk} to neural networks\cite{aslan2021cnn}, with problem formulations of both supervised and transfer learning\cite{pathak2020deep,liu2022reduce}. However, to the best of our knowledge, there has not been study that look into the long-term prediction performance model over time by the time we wrote this paper.

\section{Method} \label{sec:method}

\subsection{Upper bound of the sample variance of AUCROC}
We derive the upper bound of the sample variance from DeLong’s method for estimating the variance of AUCROC. This upper bound is used later in case there is very few positive samples in an observation window as an conservative estimation of variance of AUCROC. Consider a binary classification problem, let m denotes the number of positive samples (class 1) and n denotes the number of negative samples (class 0). Assume there are two probability density functions, $P_x$ and $P_y$, such that $P_x$ represents the distribution of predicted scores of positive samples and the other one $P_y$ represents the predicted scores of negative samples. In the following equations, $x$ is drawn from $P_x$, and $y$ is drawn from $P_y$. Then the mean AUCROC $\theta$ can be estimated by the Mann-Whitney statistics
\begin{equation}\label{eq:MW-stst}
\hat{\theta} = \frac{1}{mn}\Sigma^{m}_{i=1}\Sigma^{n}_{j=1}\mathbbm{1}_[x_{i}>y_{i}]
\end{equation}
Where $m$ is the number of sample scores $x$ drawn from $P_x$, $n$ is the number of sample scores $y$ drawn from $P_y$, $\mathbbm{1}_[x_{i}>y_{i}]$  is the characteristic function giving 1 when the condition $x_i>y_j$ is satisfied, otherwise zero. A practical working assumption is that machine learning models seldom generate the same prediction scores for two different data points. Therefore, ties are not of great concern here in the following derivations, and hence, the situation is simplified.

\begin{equation}\label{eq:1}
    V_{10}(x)=\frac{1}{n}\Sigma^{n}_{j=1}\mathbbm{1}_[x>y_{i}]
\end{equation}
\begin{equation}\label{eq:2}
V_{01}(y)=\frac{1}{m}\Sigma^{m}_{i=1}\mathbbm{1}_[x_{i}<y]
\end{equation}

\begin{equation}\label{eq:3}
    S_{10}=\frac{1}{m-1}\Sigma^{m}_{i=1}(V_{10}(x_i)-\hat{\theta})^2
\end{equation}
\begin{equation}\label{eq:4}
    S_{01}=\frac{1}{n-1}\Sigma^{n}_{j=1}(V_{01}(y_i)-\hat{\theta})^2
\end{equation}

Finally, the sample variance can be estimated using equation \ref{eq:5}. An upper bound of the variance can be derived in equation \ref{eq:6}, as both $S_{10}$ and $S_{01}$ are actually bounded by 1. To see that, $\theta$, $V_{10}(x)$,$V_{01}(y)$ are all in $[0,1]$. What's more, $s^2 \in [0,1]$, given $s\in[0,1]$. In a healthcare application of predictive models, one often faces an imbalanced dataset. We note that $\frac{1}{m}+\frac{1}{n} \approx \frac{1}{m}$,if $m \ll n$ in an imbalanced setting, meaning the positive samples contribute overwhelmingly to the sample variance.
\begin{equation}\label{eq:5}
    Var(\hat{\theta})=\frac{1}{m}S_{10}+\frac{1}{n}S_{01}
\end{equation}
\begin{equation}\label{eq:6}
    Var(\hat{\theta}) \leq \frac{1}{m}+\frac{1}{n}
\end{equation}

\subsection{Proposed framework for estimating model performance over time}
%notations
%notations
We introduce notations and symbols in the context of binary classification problems.
\begin{table}
    \centering
\begin{tabular}{rl}
 \toprule
Symbols & Meaning\\
\midrule
$\theta, \hat{\theta}$& evaluation metric of the binary classifier\\
$\Box_{t}$ & subscription denotes the time step\\
$m_t$ & number of positive samples\\
$n_t$ &number of negative samples \\
$z_t,r_t$ & sample mean and sample variance of the \\
& evaluation metric at current time t\\
$p_{t,t-1}$ & variance/covariance extrapolated \\
& based on estimation from time t-1\\
$p_{t,t}$ & current variance/covariance estimation\\
$K_t$& Kalman Gain\\

\bottomrule
\end{tabular}
        \caption{Notations and Symbols} 
        \label{tab:Notations}
\end{table}

\begin{figure*}[t]
  \centering
  \includegraphics[width=14cm]{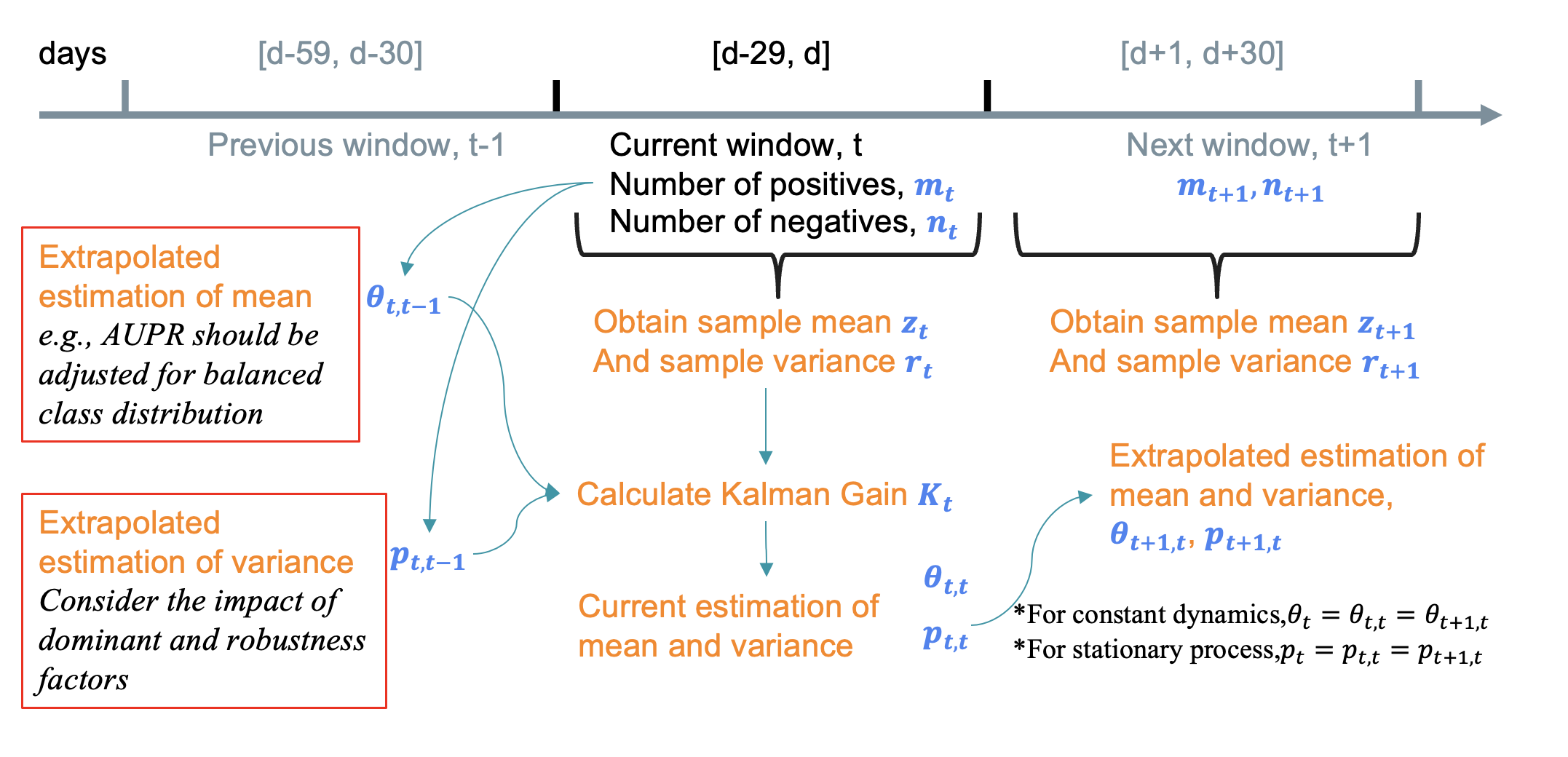}
  \caption{\centering Proposed Kalman Filter based framework for estimating model performance over time. Changes made to the classical Kalman filter is highlighted in the red boxes.}
  \label{fig:KFarchi}
\end{figure*}

\begin{table}[h]
    \centering
    \begin{tabular}{ll}
    \toprule
       Steps  &  Formula\\
    \midrule
        1 & $z_t$, \\
        & if $m_t>$ threshold, use $r_t$\\
        & else use $\frac{1}{m_t}+\frac{1}{n_t}$\\
        2 & $p_{t,t-1}=\frac{1}{m_t}S_{10,t-1}+\frac{1}{n_t}S_{01,t-1}$ \\
        3 & $K_t=\frac{p_{t,t-1}}{p_{t,t-1}+r_t}$ \\
        4 & $\theta_t = \theta_{t-1}+K_t(z_t-\theta_{t-1})$\\
        5 & $p_{t,t}=(1-K_t)p_{t,t-1}$\\
         \bottomrule
    \end{tabular}
    \caption{Kalman filter steps for AUCROC}
    \label{tab:KFsteps}
\end{table}

We describe the steps per time step iteration in table \ref{tab:KFsteps}. This algorithm is designed for AUCROC, therefore we use $z_t$ to denote the sample AUCROC at time $t$ specifically. For other model performance metrics, adjustments are needed accordingly. In the first step, sample mean AUCROC $z_t$ is calculated using all predictions made during the current window. Sample variance $r_t$ is estimated by DeLong’s method described in the previous section. Since we are expecting an imbalanced data batch per time window, if there is not enough positive samples, we conservatively use the upper bound of sample variance. Then, the previous variance estimate  $p_{t,t}$ is extrapolated to $p_{t,t-1}$, following DeLong’s equation \ref{eq:5}. 

\begin{enumerate}
    \item Estimate mean $z_t$ and sample variance $r_t$ of the performance metric based on the data in the moving window which ends at time t. Since we are expecting an imbalanced data batch, if there is not enough positive samples, we conservatively use the upper bound of sample variance.
    \item Extrapolate variance according to the number of positive $m_t$  and negative samples $n_t$ at current time step. 
    \item Calculate Kalman Gain $K_t$ at time t using sample variance $r_t$ and the extrapolated estimation variance $p_{t,t-1}$
    \item Obtain the filtered value of performance metric.
    \item Update variance. Equivalently, ($1-K_t$) is applied to $S_{10}$,$S_{01}$ at time t. $p_{t,t}$ is used for constructing a 95\% confidence interval.
\end{enumerate}

\section{Experiments}  \label{sec:exp}
\subsection{Synthetic Data}
To demonstrate that the proposed framework is still able to detect change of performance in time, we have designed a synthetic simulation dataset which consist of three phases, changes in both sample sizes and positive ratio/class distribution, shifts in model performance and 60 time steps in total. Two simulate the change of performance, we specify the distribution of scores of positive samples and scores of negative samples by two independent normal distributions 
In \textbf{Phase 1}:Step 0-19. Ground truth AUCROC and (Binary)Class distribution (5\%) stays the same. However, the total number of samples starts at 5000 and gradually decreased to 50. 
In \textbf{Phase 2}:Step 20-39. Ground truth AUCROC is unchanged. The total number of samples remains 400 during this phase. However, positive ratio is gradually decreased to 2\%. 
In \textbf{Phase 3}: Step 40-59. Declined AUCROC, same positive ratio (2\%) as in the previous phase, gradually increasing total number of samples from 400 to 5000. 

Python implementation can be found in the supplementary material.

\begin{figure}[h]
  \centering
  \includegraphics[width=\linewidth]{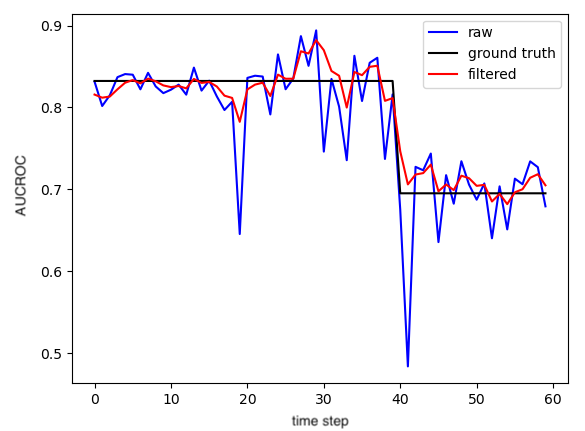}
  \caption{Results of a 3-Phase Simulation. Ground Truth AUCROC is in black, raw AUCROC is in blue and filtered AUCROC is in red. \textbf{Phase 1}:Step 0-19. Ground truth AUCROC and (Binary)Class distribution (5\%) stays the same. However, the total number of samples starts at 5000 and gradually decreased to 50. \textbf{Phase 2}:Step 20-39. Ground truth AUCROC is unchanged. The total number of samples remains 400 during this phase. However, positive ratio is gradually decreased to 2\%. \textbf{Phase 3}: Step 40-59. Declined AUCROC, same positive ratio (2\%) as in the previous phase, gradually increasing total number of samples from 400 to 5000.}
  \label{fig:simres}
  
\end{figure}

The result is shown in Figure \ref{fig:simres}, where the black line denotes the ground truth AUCROC, the blue line denotes the AUCROC calculated base on data from the current window and the red line represent the filtered AUCROC.

\subsection{Real Dataset}
We start from data processing and model training steps. Then the proposed method is applied.
\subsubsection{Data Processing}
Baseline variables and demographics are collected at the time of admission. These includes initial readings of vital signs, age, gender, race, ethnicity, pregnant flag, and smoking status, umber of previous hospital stays and number of previous ICU stays. No vaccine information was recorded by the Abbott Northwestern Hospital’s Enterprise Data Warehouse (EDW) during the period of data collection from 2020 June to March 2023. We have determined the cut-off admission date to be December 1st, 2022, so that all patients included the dataset have a known end of stay outcome. Vital signs such as systolic/diastolic blood pressure, heart rate, respiration rate, saturation of oxygen($SpO_2$) and body temperature are collected multiple times a day for every patient, regardless of their Intensive Care Unit(ICU) status. Patient daily feature vectors are then derived from all readings of vitals measured in that day. Each vital sign is aggregated to mean, median, min, max, trend(slope), detrended variation and the count of measurements per day. All vital sign features(mean, median, min, max) except temperatures are log transformed following Yeo-Johnson's method\cite{weisberg2001yeo}. We take $1-SpO_2$ before log transformation, assuming $SpO_2$ is between 0 and 1. Body temperatures are standardized to z-score of a standard normal distribution. To model the temporal correlation, we also add the first order difference of all the variables in a daily feature vector. Empirically, adding higher order differences does not significantly improve performance of 5-fold cross validation on the training set. Please refer to the supplementary materials for 

\subsubsection{Mortality Prediction Model Training}

We train a XGBoost\cite{chen2016xgboost} classifier and tune model parameters on the data collected till Dec 15, 2020, and tested retrospectively on the test set collected onward till the end of 2022. Model parameters are determined using 5-fold cross validation on the training set. Learning rate is set to 0.05, L1 regularization 0.01, no L2 regularization and maximum depth to be 3. Isotonic regression is tested as a method of score calibration. However, empirically, we have found that the performance, namely AUCROC, of probability calibrated classifiers to significantly worse than uncalibrated scores for some reasons. Outputs of the model are then grouped by dates and ranked descendingly by their predicted scores. A daily report is generated, highlighting the patients above the threshold, the history of past predictions and the current feature importance. True positive predictions, false positive predictions and false negative predictions are randomly drawn from the test set.
\subsubsection{Filtered AUCROC}
\begin{figure*}[t]
  \centering
  \includegraphics[width=14cm]{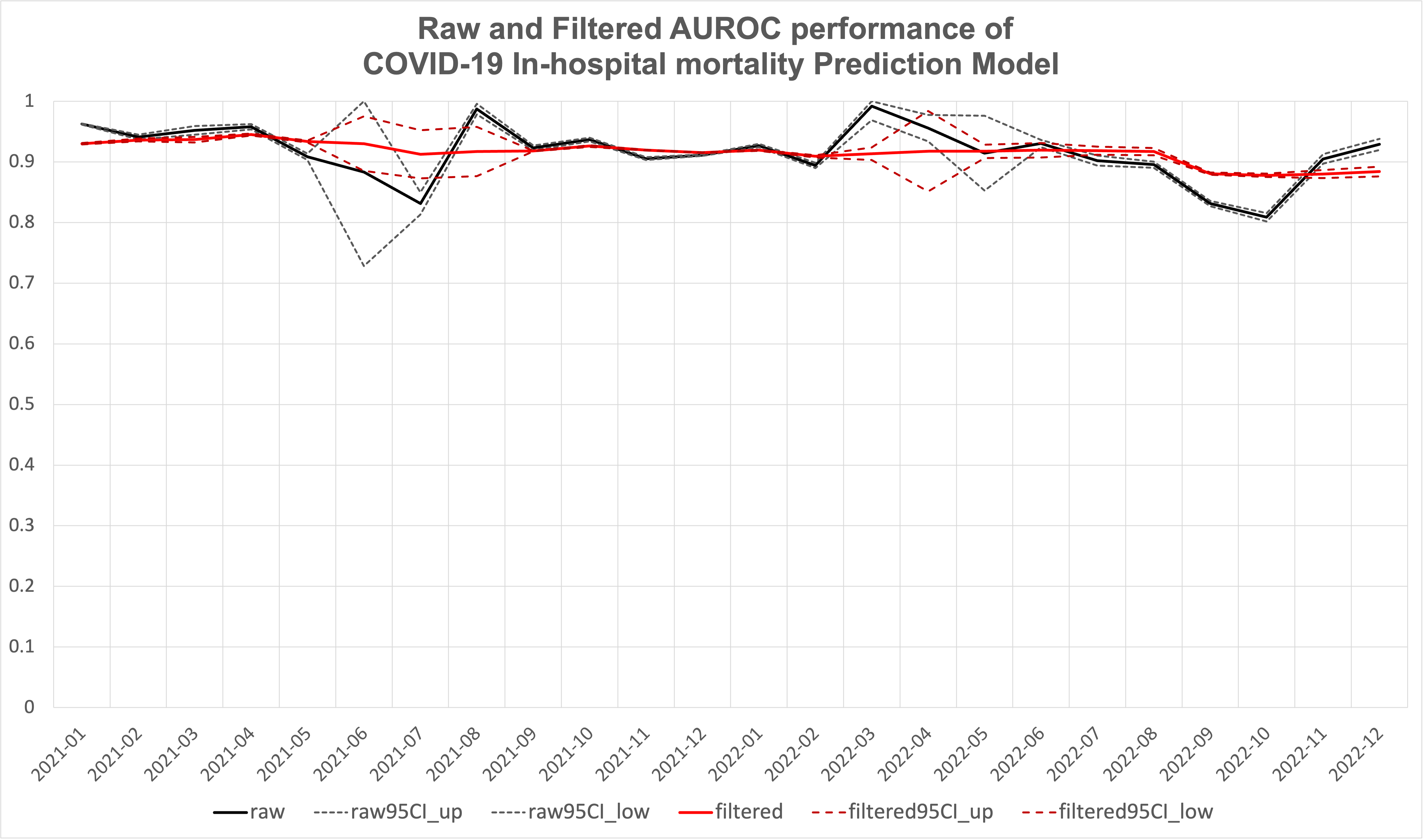}
  \caption{\centering Raw monthly AUCROC values of 2 days ahead in-hospital mortality prediction model for COVID-19 patients are represented by the black solid line, the same as the one in Fig. 2.  Filtered AUCROC in red. 95\% confidence intervals of both raw and filtered AUCROC are in dotted/dashed lines of the respective color.}
  \label{fig:results}
\end{figure*}

Now we are ready to answer the questions raised in section \ref{sec:2} and Figure. \ref{fig:monthlyAUCROC}. In Figure \ref{fig:results}, Raw monthly AUCROC values of 2 days ahead in-hospital mortality prediction model for COVID-19 patients are represented by the black solid line. Filtered AUCROC in red. 95\% confidence intervals\cite{cortes2004confidence} of both raw and filtered AUCROC are in dotted/dashed lines of the respective color. The result suggests that the performance of the model remains stable through 2021 and 2022 despite the changes in class distributions, number of patients, length of stay and root causes such as evolving virus variants and improving treatment medications and guidelines.

\begin{table*}[h]
    \centering
\begin{tabular}{rrrrrrr}
 \toprule
month   & raw    & raw95CI\_up & raw95CI\_low & filtered & filtered95CI\_up & filtered95CI\_low \\
\hline
\midrule
2021-01 & 0.9624 & 0.9635      & 0.9613       & 0.9300   & 0.9311           & 0.9289            \\
2021-02 & 0.9410 & 0.9449      & 0.9371       & 0.9362   & 0.9384           & 0.9340            \\
2021-03 & 0.9520 & 0.9592      & 0.9448       & 0.9368   & 0.9415           & 0.9321            \\
2021-04 & 0.9581 & 0.9626      & 0.9536       & 0.9452   & 0.9470           & 0.9434            \\
2021-05 & 0.9084 & 0.9136      & 0.9032       & 0.9336   & 0.9352           & 0.9320            \\
2021-06 & 0.8834 & 1.0000      & 0.7286       & 0.9302   & 0.9751           & 0.8853            \\
2021-07 & 0.8316 & 0.8498      & 0.8134       & 0.9127   & 0.9524           & 0.8730            \\
2021-08 & 0.9872 & 0.9960      & 0.9784       & 0.9173   & 0.9579           & 0.8767            \\
2021-09 & 0.9231 & 0.9270      & 0.9192       & 0.9180   & 0.9185           & 0.9175            \\
2021-10 & 0.9369 & 0.9399      & 0.9339       & 0.9262   & 0.9275           & 0.9249            \\
2021-11 & 0.9054 & 0.9077      & 0.9031       & 0.9194   & 0.9201           & 0.9187            \\
2021-12 & 0.9113 & 0.9125      & 0.9101       & 0.9155   & 0.9161           & 0.9149            \\
2022-01 & 0.9269 & 0.9296      & 0.9242       & 0.9193   & 0.9202           & 0.9184            \\
2022-02 & 0.8943 & 0.8985      & 0.8901       & 0.9092   & 0.9109           & 0.9075            \\
2022-03 & 0.9922 & 1.0000      & 0.9687       & 0.9135   & 0.9238           & 0.9032            \\
2022-04 & 0.9559 & 0.9777      & 0.9341       & 0.9177   & 0.9841           & 0.8513            \\
2022-05 & 0.9145 & 0.9763      & 0.8527       & 0.9176   & 0.9286           & 0.9066            \\
2022-06 & 0.9304 & 0.9364      & 0.9244       & 0.9193   & 0.9312           & 0.9074            \\
2022-07 & 0.9022 & 0.9106      & 0.8938       & 0.9184   & 0.9251           & 0.9117            \\
2022-08 & 0.8954 & 0.9005      & 0.8903       & 0.9170   & 0.9230           & 0.9110            \\
2022-09 & 0.8313 & 0.8356      & 0.8270       & 0.8805   & 0.8823           & 0.8787            \\
2022-10 & 0.8087 & 0.8155      & 0.8019       & 0.8779   & 0.8807           & 0.8751            \\
2022-11 & 0.9050 & 0.9130      & 0.8970       & 0.8800   & 0.8865           & 0.8735            \\
2022-12 & 0.9289 & 0.9380      & 0.9198       & 0.8840   & 0.8921           & 0.8759            \\       
 \bottomrule
\end{tabular}
        \caption{Raw and Filtered AUCROC} 
        \label{tab:kfres}
\end{table*}

\section{Discussion}\label{sec:discussion}
The proposed method utilizes a conservative upper bound of variance, in some case may lead to slow adaptation. However, in the specific case of AUCROC of in-hospital mortality prediction models, since the performance data is already aggregated to a monthly level, this problem is mitigated in some sense. Another potential issue with long-term performance monitoring of predictive models is the problem of setting the p-value threshold and multiple comparisons. The problem of multiple comparisons against the confidence interval is a realistic concern when tracking model performance over time. The Kalman filter takes care of it by shrinking the estimation of variance, as step 5 in table \ref{tab:KFsteps} suggests.

\begin{figure}[t]
  \centering
  \includegraphics[width=\linewidth]{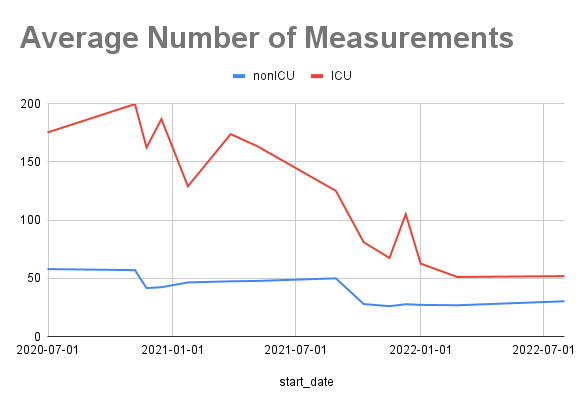}
  \caption{Measurement frequency of vital signs per patient per day. The two drops were due to changes of hospital operation plan. Namely, the per patient workload of nurses were reduced so that more patients could be taken care of. As a result, the average number of vital sign measurements per patient per day was reduced.}
  \label{fig:measurement frequency}
\end{figure}

Besides the analysis shown in the experiment section, we compared the filtered results against the change in measurement frequency per patient per day. As shown in Figure\ref{fig:measurement frequency}, there were two changes of hospital operational plan during 2020 to 2022 at Abbott Northwestern hospital. 

The hospital went into ``crisis mode'' to deal with the overwhelming number of patients and the ``burn-out'' phenomenon of physicians and nurses. The per patient workload of nurses were reduced so that more patients could be taken care of, which means the average number of vital sign measurements per patient per day was reduced. To our surprise, the model performance does not fluctuate much in response to hospital operational plan changes. Please bear in mind that the model is trained only on 2020 data and was not retrained on new batch of data. This finding may shed lights on new possibilities of reducing the cost of healthcare while maintaining the same quality of care.

For future work, we consider the filter of AUCPR. Note that while sample size and class distributions are robustness factors for AUCROC, they are dominant factors for AUCPR.

\section{Conclusion}
In this paper, we start from a question which is rooted in real-world scenario, that is, how to compare performance of the same model over different time periods. Through analysis, we identify dominant and robustness factors of evaluation metrics. Thus, a Kalman filter based framework is proposed to adjusted for the shift of class distributions and the change in sample size. Experiments on synthetic datasets demonstrate its ability to not only remove noises, but also to  track change of performance correctly. Although the problem has healthcare contexts, we believe our method is widely applicable and can be potentially adapted to other performance metrics.

\section{Acknowledgement}
The research plan and the use of patient data is approved by the IRB at Abbott Northwestern Hospital, Minneapolis, MN. This study is supported by the Abbott Northwestern Foundation (ANWF21-0102). We also sincerely thank the Minnesota Supercomputing Institute (MSI) at the University of Minnesota\footnote{\url{ http://www.msi.umn.edu}} for providing computational resources.

\end{document}